\documentclass{article}
\usepackage{spconf,amsmath,graphicx}
\usepackage[hyphens]{url}
\usepackage{amsmath}
\usepackage{paralist} 
\usepackage{booktabs}
\usepackage[dvipsnames]{xcolor}
\usepackage{microtype}
\usepackage{ragged2e}

\PassOptionsToPackage{hyphens}{url}
\usepackage{caption}
\usepackage{subcaption}
\RequirePackage{xcolor}
\usepackage{multirow}
\usepackage[hyphens]{url}
\usepackage[pagebackref=false,breaklinks=true,colorlinks,bookmarks=false]{hyperref}

\usepackage{color}

\def\ie{{\em i.e.}}
\def\eg{{\em e.g.}}

\def\eg{\emph{e.g.}}

\def\ie{\emph{i.e.}}

\usepackage{enumitem}

\begin{document}

\title{Detection of Real-time DeepFakes in Video Conferencing with Active Probing and Corneal Reflection}

\name{Hui Guo, Xin Wang, Siwei Lyu}

\address{Department of Computer Science and Engineering\\ University at Buffalo, State University of New York, USA. \\
{\tt\{hguo8,xwang264,siweilyu\}@buffalo.edu}  }

\maketitle

\begin{abstract}
The COVID pandemic has led to the wide adoption of online video calls in recent years. However, the increasing reliance on video calls provides opportunities for new impersonation attacks by fraudsters using the advanced real-time DeepFakes. Real-time DeepFakes pose new challenges to detection methods, which have to run in real-time as a video call is ongoing. In this paper, we describe a new active forensic method to detect real-time DeepFakes. Specifically, we authenticate video calls by displaying a distinct pattern on the screen and using the corneal reflection extracted from the images of the call participant’s face. This pattern can be induced by a call participant displaying on a shared screen or directly integrated into the video-call client. In either case, no specialized imaging or lighting hardware is required. Through large-scale simulations, we evaluate the reliability of this approach under a range in a variety of real-world imaging scenarios. 
\end{abstract}

\begin{keywords}
Real-time DeepFake, Corneal Reflection
\end{keywords}


\section{Introduction}
\vspace{-0.2cm}

Video calls are increasingly replacing in-person meetings and phone calls in recent years, mainly due to the high demand of remote working during the COVID pandemic. For instance, at the end of 2019, the Zoom video conferencing platform had only about 10 million users. By late April of 2021, that figure had surged to over 200 million, a 20-fold increase. However, the wide adoption of video calls as a means of meeting and inter-person communication has also given rise to new forms of deception. In particular, the lack of physical presence opens the gate for digital impersonation in video calls using DeepFakes (\ie, AI-synthesized human face videos). The most recent tools (\eg, Avartarify \cite{avatarify} and DeepFaceLive \cite{DeepFaceLive}) have enabled the synthesis of DeepFakes in real-time and piped through a virtual camera. The DeepFakes are either in the form of face-swap or face puppetry \cite{lyu_icmew20}. Although there are still artifacts in the real-time DeepFakes \cite{futurism}, the continuing improvement of the synthesis technology means that it will become increasingly difficult to distinguish a real person from an AI-synthesized person at the other end of a video call. Indeed, recent years are seeing such frauds emerged with an alarming speed and start causing real damage \cite{Avinteractive}. 

\begin{figure}[t]
\centering
\includegraphics[width=.47\textwidth]{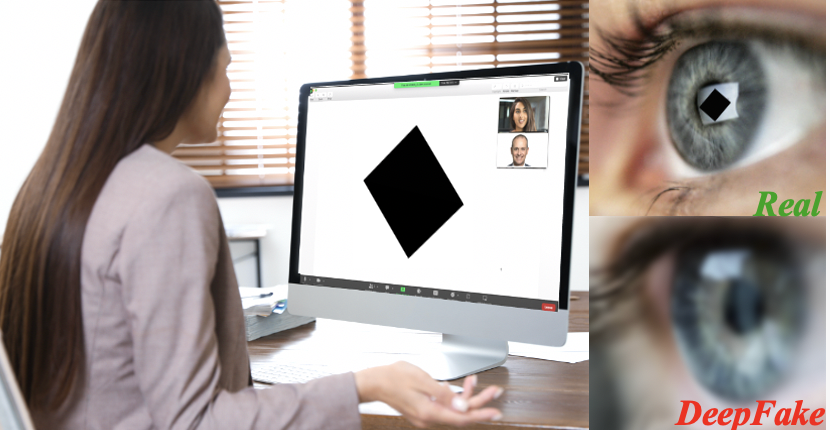}
\vspace{-1em}\caption{\em \textbf{Left:} A video call attendant is being actively authenticated with the live patterns shown on the screen.  \textbf{Right:} A real person's cornea will produce an image of the pattern shown on the screen, while a real-time DeepFake cannot. The figures are for demonstration. For actual results, see Fig. \ref{fig_experiment}. }
\label{fig_introduction}
\vspace{-2em}
\end{figure}

The real-time DeepFakes pose new challenges to existing detection methods, which are mostly {\em passive}, in that they classify an input video into the category of authentic or DeepFake. Most of these methods struggle to achieve the levels of accuracy that would be needed to be incorporated into a practical video-conferencing application and run in real-time. On the other hand, new approaches based on {\em active} forensics, which interfere with the generation process to improve detection efficiency and accuracy, \eg, \cite{sun_etal_wifs20,Gerstner_2022_CVPR}, are gaining momentum recently. In particular, the work of \cite{Gerstner_2022_CVPR} exploits the unique constrained environment afforded by a video-conferencing call to detect real-time DeepFakes by varying the lighting pattern on screen and extracting the same lighting variation from the attendant's face. As the current real-time DeepFake synthesis methods are not sufficiently adaptable to capture such subtle changes, the lack of consistent lighting variation can be used as a telltale sign of synthesis and impersonation. However, controlling and estimating the subtle change of screen lighting may not be reliable as it can be affected by other environmental factors, such as the ambient light, room setting, and makeup.

In this work, we describe a new active forensic approach to exposing real-time DeepFakes. The main idea is illustrated in Fig. \ref{fig_introduction}. This method can be initiated by a call participant or directly integrated into the video-call client\footnote{We assume that a consensus can be obtained from the attendants to use their imagery for authentication purposes without privacy issues. This would be the same agreement required when the video call is live recorded.}. First, we  briefly display  a distinct pattern, which will be referred to as the {\em probing pattern}, on the shared screen during an ongoing video call. The image of the attendant's face will be captured by the camera, and we will focus on the cornea areas. As the attendant sits before the camera in a video call and the human cornea is mirror-like and highly reflective, the probing pattern on the screen should leave a reflective image on the cornea that can subsequently be extracted from the face image and compared with the probing pattern. We provide an automatic pipeline to display the probing pattern, capture the face image, extract the cornea reflections, and compare with the original probing pattern. Our experiments with several state-of-the-art real-time DeepFake synthesis models show that they cannot recreate the probing pattern at the synthesized cornea region at all in a variety of real-world settings. Compared with the work in \cite{Gerstner_2022_CVPR}, our active detection method is less limited by the lighting environment. In addition, our method does not rely on complicated trained models, which allows use in a real-time video conferencing environment easily. On the other hand, our method can reliably extract and compare probing patterns to authenticate real persons under a range of imaging scenarios and validate this approach. 

\section{Related Works}
\vspace{-0.2cm}



\begin{figure}[t]
\centering
\includegraphics[width=.48\textwidth]{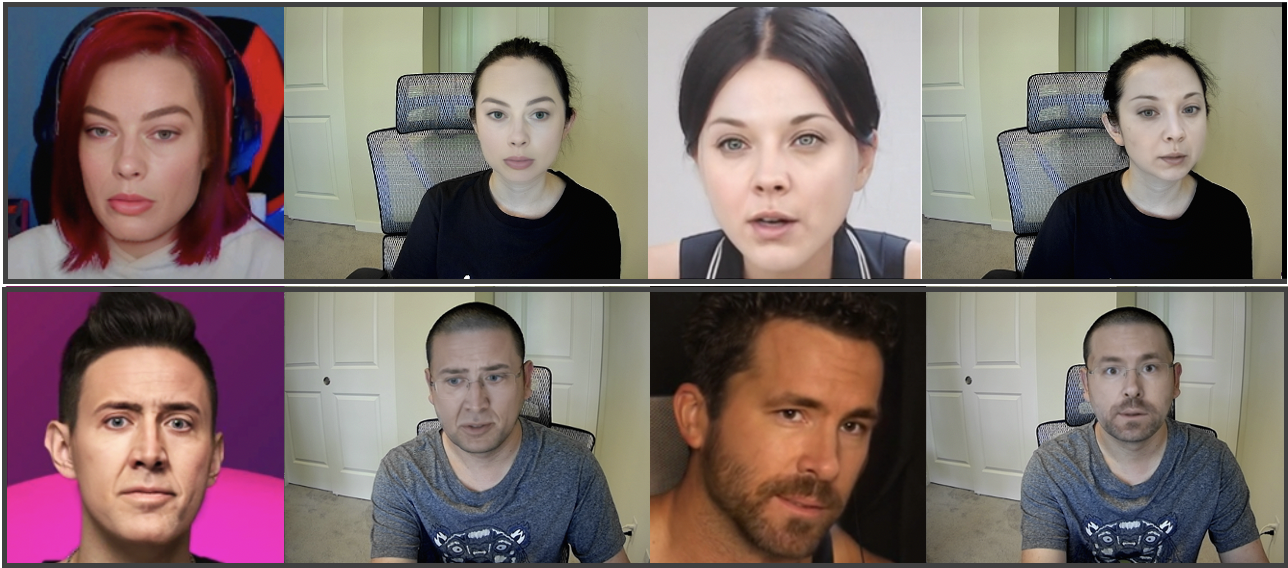}
\vspace{-2em}
\caption{\em Examples of video-conferencing DeepFake using DeepFaceLive \cite{DeepFaceLive}. For each pair, \textbf{Left:} The template Faces, \textbf{Right:} The DeepFakes. }
\vspace{-1em}
\label{fig_DeepFaceLive}
\end{figure}

\noindent\textbf{Real-time DeepFake Synthesis.} 
DeepFakes are made for real-time synthesis in recent years.
DeepFaceLive \cite{DeepFaceLive} was proposed to DeepFake in the real video-conferencing scenario. It obtains higher visual quality and real-time speed that could be used in practice. 
Using the DeepFaceLive, the users can swap their faces from a real webcam using trained face-swapping models in real-time. The generated fake screen in the DeepFaceLive software can be passed to the video-conferencing software via virtual camera software (\eg, OBS-VirtualCam \cite{DeepFaceLive}). For example, in the Zoom software \cite{zoom}, the host can select to use a virtual camera instead of the actual camera to display the fake screen from the DeepFaceLive in the Zoom meeting.
Examples of running DeepFaceLive in a Zoom meeting are shown in Fig. \ref{fig_DeepFaceLive}.





\noindent\textbf{DeepFake Detection Using Eye Biometrics.} Biometric cues from the eyes have been used for the detection of GAN-generated still images \cite{matern2019exploiting,hu2021exposing,guo2022eyes,guo2021robust,wang2022gan,guo2022open}. 
The work \cite{hu2021exposing} uses the inconsistency of corneal specular highlights in the two eyes to identify AI-synthesised faces. 
More recently, the work \cite{guo2022eyes} spot the AI-synthesised faces by detecting the inconsistency of pupil shapes. 
These methods are further extended in \cite{guo2021robust} by using an attention-based robust deep network, where the inconsistent components and artifacts in the iris region of the GAN-face are highlighted in the attention maps clearly. 
Although effective in exposing GAN-generated faces in high-resolution still images in a passive setting, these methods may not work to catch real-time DeepFake videos that are used in video conferences. 


\noindent\textbf{Active Detection of DeepFakes.} The active detection for DeepFakes differs from the existing detection methods \cite{pu2022learning} in that it interferes with the generation process to make detection easier. Early work in \cite{sun_etal_wifs20} obstructs the DeepFake generation by attacking a key step of the generation pipeline, \ie, facial landmark extraction. The method generates adversarial perturbations \cite{hu2021tkml} to disrupt the facial landmark extraction, such that the DeepFake models cannot locate the real face to swap. Active illumination artifacts are studied for exploring the DeepFakes. For example, the work \cite{shang2020protecting} shows that the correspondence of brightness of the facial appearance in different active illumination can be used as a signal for active DeepFakes detection. Motivated by this work, \cite{Gerstner_2022_CVPR} proposed a new active method for video-conferencing DeepFakes detection using active illumination. 


\begin{figure*}[t]
\centering
\includegraphics[width=.99\textwidth]{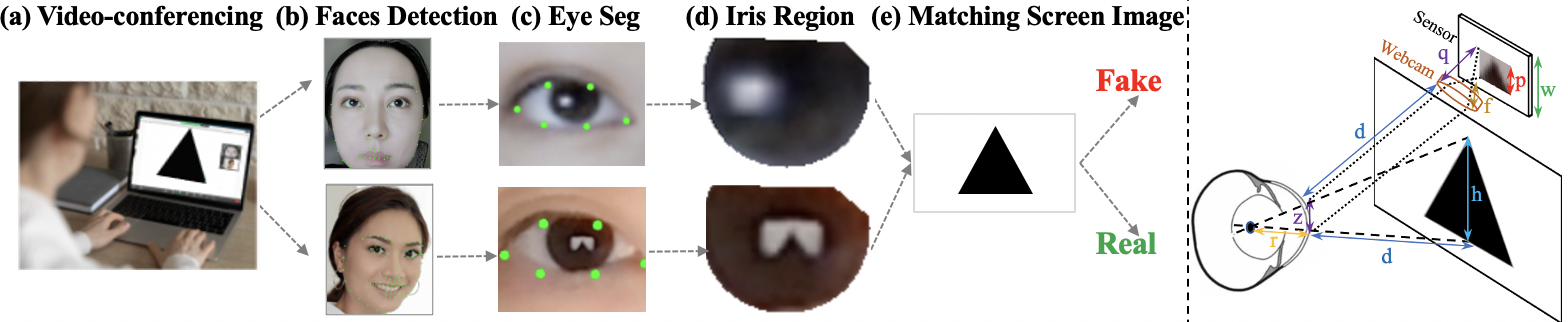}
\vspace{-1.1em}
\caption{\em \textbf{Left:} The overall process of the proposed method. See texts for details. \textbf{Right:} Visual model to estimate the size of probing pattern when it is shown on the sensor, i.e. variable $p$. We assume the probing pattern is square, and also estimate how many pixels in the area of $p^2$ on the sensor.}
\vspace{-1em}
\label{fig_overview}
\end{figure*}

\section{Method}

The overall process of our method is shown in Fig.~\ref{fig_overview}. In a standard video conference setting, a person sits in front of a laptop computer, and her eyes are captured by the webcam, Fig. \ref{fig_overview} (a). To verify if the attendant(s) is indeed a real person instead of a synthesis from real-time DeepFake models, the host will briefly put up the {\em probing pattern} on the shared screen. The probing pattern is a simple geometric shape on a white background to have good contrast. We expect the real attendants' eyes have reflections of the {\em probing pattern}, while a real-time DeepFake will not. We first capture an image of the attendant's face and then run a face detector and extract facial landmarks using Dlib~\cite{king2009dlib}, Fig. \ref{fig_overview} (b). From the facial landmarks, we localize the eye region, Fig. \ref{fig_overview} (c), and then segment out the iris part using an edge detector and the Hough transform as in \cite{hu2021exposing}, Fig. \ref{fig_overview} (d). 
The segmented iris images are then passed to the template matching steps for automatic DeepFake detection, Fig. \ref{fig_overview} (e), where we compare the corneal reflection with the probing pattern. The matching of the two indicates a real person, and the lack of matching suggests a possible real-time DeepFake impersonation. 

Our method is based on the assumption that a probing pattern on the screen in a video conference can be reliably captured. A key question is if we can have a sufficiently large image from the corneal reflection to match the probing pattern. In the following, we will give an estimation of the number of pixels using an idealized model of a real video conferencing scenario, Fig. \ref{fig_overview} (Right). 


Assuming the probing pattern is symmetric in both directions with size $h$ (in centimeters). For simplicity, we only consider the vertical dimension. For a laptop display of dimension $30.41 \times 21.24$ centimeters (cms), we choose $h \approx 14.5 $ cms, which account for 70\% of the height on the display. The attendant is assumed to sit at $d = 30$ cms away from the screen, which is approximately the distance from the center of the probing pattern to the center of the eye. We further denote $r$ as the radius of the eyeball, which is approximate $1.25$cm for a healthy adult \cite{Human_eye}. The built-in webcam of the laptop computer has a sensor of height $w = 45 \times 10^{-2}$ cms. A vertical scan line of the sensor has $M = 720$ pixels. We also assume the focal length of the webcam is set to $f = 50 \times 10^{-2}$ cms. 

We first compute the vertical height of the corneal reflection of the probing pattern (in cms), which is denoted as $z$. With a simple geometric relation, we have $\frac{z}{r} \approx \frac{h}{d}  \Longrightarrow    z = {hr \over d}$.
Next, using the focal length formula \cite{Focallength}, we can estimate the horizontal distance from the lens of the camera to the sensor, $q$, as
${1 \over d} + {1 \over q} = {1 \over f} \Longrightarrow q = {df \over d-f}$.
With the geometrical relations between $z$, $d$, and $q$, we can have an estimation of the sensor image of the corneal reflection $p$, as
${p \over q} = {z \over d} \Longrightarrow p = {qz \over d} = {hrf \over d(d -f)}$.
This corresponds to a total number of ${pM \over w}$ pixels in the vertical direction for the sensor image of the corneal reflection. As the last step, assuming symmetry between the vertical and horizontal directions again, the number of pixels of the captured corneal reflections is $\left({hrfM \over wd(d -f)}\right)^2$. Plugging actual numbers for these variables, we can obtain the actual number of pixels of the corneal reflection. With the previous setting, the corneal reflection will have roughly 256 pixels in the image. 

\begin{figure}[t]
\centering
\includegraphics[width=.48\textwidth]{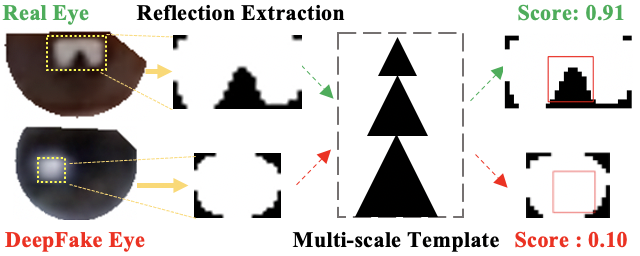}
\vspace{-2em}
\caption{\em Overview of the probing pattern matching. The multi-scale templates are generated using the scaled probing pattern. }
\vspace{-1em}
\label{fig_match}
\end{figure}

The previous derivation establishes the approximate size of the corneal reflection image of the probing pattern, the detection of real-time DeepFake impersonation in a video conference could be a straightforward search in the image of the probing pattern. However, there are some subtleties that motivate a more elaborated solution.  First, because of the ambient light in the surrounding environment and the gamut difference between the screen and camera, directly comparing RGB images could lead to inaccurate matching. In Fig. \ref{fig_colors} of Section \ref{sec_exp} we experimentally demonstrate the effects of color to the matching performance. Since the shape of the probing pattern is more essential in this case, we binarize the patterns obtained from the corneal reflection images and use the binarized mask to compare with the probing pattern, Fig. \ref{fig_match}. This is because we can use probing patterns with simple shapes and high contrast (\eg, a triangle shape of saturated color on white background). The high contrast makes the binarization process easier, and we can use automatic thresholding algorithms \cite{hu2021exposing} for that purpose.

In addition, to proceed with the matching, we need to make the probing pattern to have a similar size as the pattern in the corneal reflection image. This can be done using the analysis of the approximate size of the shape, as in the previous section. However, we need to adjust the matching scale over a range because (i) the estimation is only approximate and (ii) in practice, the attendant's face may move and lead to different distances to the camera and display. Therefore, we generate multi-scale templates and then search the templates in the iris images to identify the occurrence of the template in the image. To generate the multi-scale templates, we need to estimate the shown size of the probing pattern in the sensor (i.e. the reflected probing pattern) as the scaling parameter. The estimated pixel number $N$ as the scaling parameter, we can generate appropriate multi-scale probing patterns as templates for template matching. For each single matching step with one of the multi-scale templates, we use the method of normalized cross-correlation (NCC) \cite{yoo2009fast}. Formally, we denote $I$ as the reflection image (See the second column in Fig. \ref{fig_match}), we search a single template $t$ (a scaled probing pattern) over $(x, y)$ under the window positioned at $(u,v)$ in the image $I$, the corresponding NCC score is calculated as follow,
\begin{equation*}
    \begin{aligned}
      & NCC(u,v) =  \\ 
      & \frac{\sum_{x,y}[I(x,y)-\bar{I}_{u,v}][t(x-u)(y-v)-\bar{t}~]}{\bigg ( \sum_{x,y}[I(x,y)-\bar{I}_{u,v}]^2\sum_{x,y}[t(x-u)(y-v)-\bar{t}~]^2 \bigg )^{\frac{1}{2}}}
    \end{aligned}
\end{equation*}

where $ \bar{t} $ is the mean of the template and $ \bar{I}_{u,v} $ is the mean of $I(x,y)$ in the region under the template. Note that, we only return a single match with the highest NCC score. 
We use the NCC implementation in the scikit-image library \cite{van2014scikit}, the red boxes in Fig. \ref{fig_match} indicate the predicted template location. As the last step, we identify videos as possible real-time DeepFakes if it has NCC lower than a preset threshold.

\begin{figure}[t]
\centering
\includegraphics[width=0.48\textwidth]{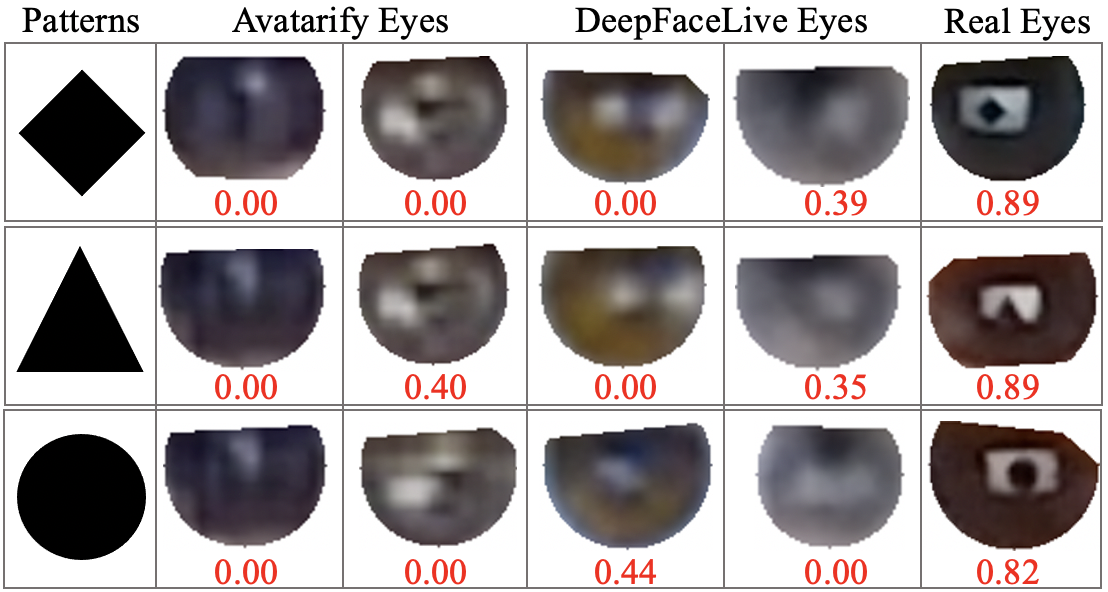}
\vspace{-2em}
\caption{\em  \textbf{Left:} The probing patterns used in our method. \textbf{Mid:} Eye reflection results using Avatar and DeepFaceLive.  \textbf{Right:}  Eye reflection results of real eyes. Due to the reflection, the probing patterns are clearly shown in the real eyes. But we can not find any probing patterns in the DeepFake eyes.}
\vspace{-1em}
\label{fig_experiment}
\end{figure}



\noindent{\bf Limitations and Counter Measures}. The effectiveness and robustness of our approach hinge on reliable detection and segmentation of face, eye/iris, and corneal reflections. Several factors may influence the final performance, such as occlusions, resolution requirements of the video-conferencing, etc. But as these components are actively studied as general Computer Vision topics, our approach will benefit from the improved techniques. 
If we use a fixed probing pattern, then a knowing adversary could predict the probing pattern and intentionally add it to the generated video with minimal temporal delay. We can counter this attack by randomizing the probing pattern and using more complex probing patterns beyond simple geometric shapes, for instance, texts representing the date and time of the meeting, so it is difficult to predict.   


\section{Results}
\label{sec_exp}
\vspace{-0.2cm}



The efficacy of our method is evaluated on two datasets. The first includes video-conferencing videos of real attendants and their DeepFake synthesis. This dataset validates our method in a realistic video-conferencing scenario. The second simulated dataset allows us to evaluate our method across a broad range of assumptions and environmental conditions.

Our real-world dataset was recorded from two users in a range of different environments. Users have placed approximately 30 cm away from the display and camera, with the probing patterns ranging in colors and shapes, such as simple diamond shapes with different colors. We use Zoom \cite{zoom} as the video-conferencing environment with default settings. 

From the real videos, we generate DeepFakes created using Avatarify \cite{avatarify}, and DeepFaceLive \cite{DeepFaceLive}. The qualitative results are shown in Fig. \ref{fig_experiment}, in which we test some geometrical signs with different shapes for the probing pattern. From the real videos, we can reliably extract the corneal reflections and match them with the input probing patterns. On the other hand, the DeepFakes do not incorporate environmental lighting and are therefore easily identifiable because in the presence of our active probing patterns, their corneal reflection patterns have a nearly zero correlation to the probing patterns. Our current method with unoptimized code has a running time of one frame per 4 seconds. It is certainly possible to improve the overall running efficiency of the algorithm by optimizing the code to be used in the video conferencing tools.


\begin{figure}[t]
\centering
\includegraphics[width=.48\textwidth]{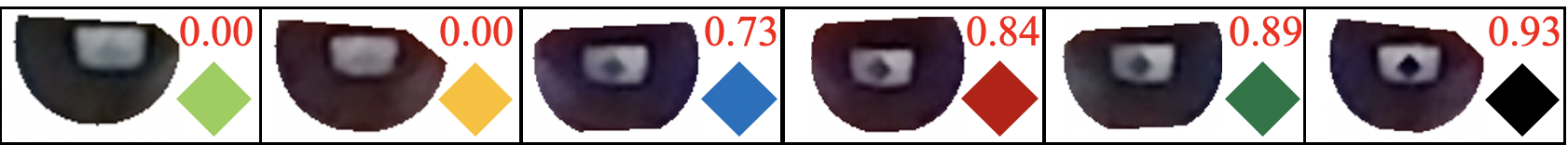}
\vspace{-2em}
\caption{\em Evaluate the effectiveness of the colors of the probing pattern where the illumination is fixed.  \textbf{Left:} Eye reflections. We can find that stronger global contrast of the sign image indicates better reflection on the iris of the real eyes. \textbf{Right:} A probing pattern with different colors. The red numbers are the corresponding NCC scores.}
\vspace{-1.5em}
\label{fig_colors}
\end{figure}

We evaluate aspects of the probing pattern and their influence on the efficacy of our approach. 

\noindent{\it Shape}. We try several different shapes for the probing screen image, as shown in Fig. \ref{fig_experiment}. We can find that these shapes can be reflected successfully in real eyes.

\noindent{\it Color and Contrast}. To understand what types of screen patterns are more effective, we synthesize common patterns with different colors for further evaluation. 
As shown in Fig. \ref{fig_colors}, we try different colors from light to dark for the probing pattern. We find that higher global contrast of the probing pattern indicates better reflection on the iris of the real eyes. Moreover, we can also see that the colors of the probing pattern are hard to show on the reflected iris. The reason may be due to the low video frame quality \cite{xu2020c3dvqa} of the video-conferencing software that uses video compressing technologies \cite{li2021deep, li2022hybrid} during the video-conferencing meeting.

\begin{figure}[t]
\centering
\includegraphics[width=.48\textwidth]{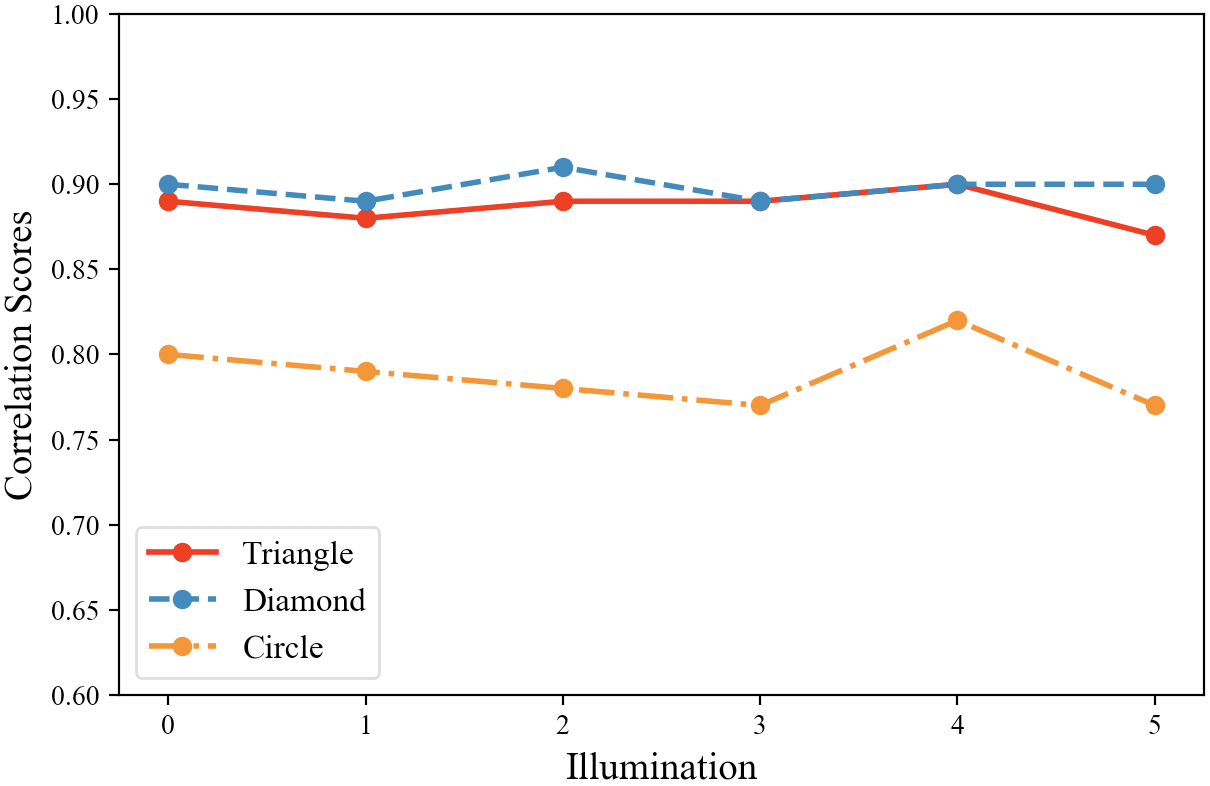}
\vspace{-1em}
\caption{\em Evaluate the influence of the illumination of indoor to our method, we change indoor light from  dark to bright. The numbers on $x$ axis indicate the light strengthens of a indoor desk lamp, where 0 indicates no indoor light, 5 indicates the maximum light strengthen 460 LM \cite{LumenUnit}.}
\label{fig_illumination}
\vspace{-1em}
\end{figure}

\noindent{\it Influence of Ambient Light.} We validate the effectiveness of our method in different illumination situations. 
We analyze the influence of the indoor light source in the experiments. We exclude all the outdoor light sources and change the indoor light from dark to bright. The results are shown in Fig. \ref{fig_illumination}. 
We find that the indoor light has a limited influence on our method because we use a very high-contrast pattern, so ambient light does not matter much. 

\section{Conclusion}
\vspace{-0.2cm}

Real-time DeepFakes pose new challenges to detection methods, which have to run in real-time as a video call is ongoing. In this paper, we describe a new active forensic method to detect real-time DeepFakes by displaying a distinct pattern on the screen and using the corneal reflection extracted from the images of the call participant’s face. 
The direction of the biometric-based active forensic approach provides a promising alternative to the widely used passive forensic methods for DeepFake detection. With the increasing quality and ubiquity of synthetic media, \cite{mentzer2022vct,wang2021one} (\eg, online streaming video compression and VR-based metaverse), the active approach has more applications as it can also be used to expose {\em unauthorized} use of synthetic models. We plan to further explore in this direction in the future. In addition, we will further improve the performance and robustness of this method's components to make it more practical.


{\small
\bibliographystyle{IEEEbib}
{\raggedright
\bibliography{refs}
}
}


\end{document}